 \documentclass[pmlr,twocolumn,10pt]{jmlr} 

\usepackage{booktabs}
\usepackage{siunitx}

\usepackage{multirow}
\usepackage{arydshln}
\usepackage{svg}
\usepackage{booktabs}
\usepackage{soul}
\usepackage{subcaption}
\usepackage[T1]{fontenc}

\theorembodyfont{\upshape}
\theoremheaderfont{\scshape}
\theorempostheader{:}
\theoremsep{\newline}

\jmlrvolume{LEAVE UNSET}
\jmlryear{2023}
\jmlrsubmitted{LEAVE UNSET}
\jmlrpublished{LEAVE UNSET}
\jmlrworkshop{Machine Learning for Health (ML4H) 2023} 

 \title[Revisiting Fine-tuning Strategies for Self-supervised Medical Imaging Analysis]{Revisiting Fine-tuning Strategies for Self-supervised Medical Imaging Analysis}

\author{%
\Name{Muhammad Osama Khan} \Email{osama.khan@nyu.edu}\\
\addr New York University, New York, USA
\AND
\Name{Yi Fang} \Email{yfang@nyu.edu}\\
\addr Center for Artificial Intelligence and Robotics, New York University Abu Dhabi, Abu Dhabi, UAE
}

\begin{document}

\maketitle

\begin{abstract}
Despite the rapid progress in self-supervised learning (SSL), end-to-end fine-tuning still remains the dominant \textit{fine-tuning} strategy for medical imaging analysis. However, it remains unclear whether this approach is truly optimal for effectively utilizing the pre-trained knowledge, especially considering the diverse categories of SSL that capture different types of features. In this paper, we present the first comprehensive study that discovers effective \textit{fine-tuning} strategies for \textit{self-supervised} learning in \textit{medical imaging}. After developing strong contrastive and restorative SSL baselines that outperform SOTA methods across four diverse downstream tasks, we conduct an extensive fine-tuning analysis across multiple pre-training and fine-tuning datasets, as well as various fine-tuning dataset sizes. Contrary to the conventional wisdom of fine-tuning only the last few layers of a pre-trained network, we show that fine-tuning intermediate layers is more effective, with fine-tuning the second quarter (25-50\%) of the network being optimal for contrastive SSL whereas fine-tuning the third quarter (50-75\%) of the network being optimal for restorative SSL. Compared to the de-facto standard of end-to-end fine-tuning, our best fine-tuning strategy, which fine-tunes a shallower network consisting of the first three quarters (0-75\%) of the pre-trained network, yields improvements of as much as 5.48\%. Additionally, using these insights, we propose a simple yet effective method to leverage the complementary strengths of multiple SSL models, resulting in enhancements of up to 3.57\% compared to using the best model alone. Hence, our fine-tuning strategies not only enhance the performance of individual SSL models, but also enable effective utilization of the complementary strengths offered by multiple SSL models, leading to significant improvements in self-supervised medical imaging analysis.
\end{abstract}

\section{Introduction}
\label{sec:introduction}
Deep learning has made remarkable progress in various domains, including medical imaging analysis~\citep{wu2019deep,mckinney2020international}. However, its reliance on large labeled datasets for training poses challenges, especially in medical imaging where annotating such datasets is time-consuming, expensive, and requires the services of experienced doctors. To address this issue, self-supervised learning (SSL) has emerged as a promising approach to learn representations from unlabeled datasets. Unlabeled medical images are more easily accessible~\citep{azizi2022robust}, making SSL an effective method for acquiring generalizable representations that can then be fine-tuned for downstream tasks with limited labeled data.

Self-supervised learning uses various pretext tasks to learn transferable representations without requiring any manual annotations. It typically involves a two-stage process: \textit{pre-training}, which learns generalizable features from unlabeled data, and \textit{fine-tuning}, which adapts the pre-trained features to the specific downstream task (e.g., classification, segmentation, etc.). While there has been significant interest in developing better self-supervised \textit{pre-training} algorithms for medical imaging~\citep{azizi2021big,chaitanya2020contrastive,chen2019self,zhou2021models,haghighi2021transferable,zhou2021preservational}, optimally utilizing these pre-trained features has received little attention, with the default end-to-end \textit{fine-tuning} still the most common fine-tuning strategy used in several state-of-the-art (SOTA) methods~\citep{taher2022caid,haghighi2022dira,azizi2022robust}. However, it remains unclear whether this approach is truly optimal for effectively leveraging the pre-trained knowledge, especially considering the diverse categories of SSL that capture distinct types of features. In this paper, we bridge this research gap by presenting the first comprehensive study that discovers effective \textit{fine-tuning} strategies for \textit{self-supervised} learning in \textit{medical imaging}.

We first develop strong contrastive and restorative SSL baselines that outperform SOTA methods across four diverse downstream tasks. Building upon these strong baselines, we then conduct an extensive analysis of fine-tuning strategies using multiple pre-training and fine-tuning datasets, as well as various fine-tuning dataset sizes. We choose MoCo-v3~\citep{chen2021empirical} and MAE~\citep{he2022masked} as archetypes of contrastive and restorative SSL respectively, utilizing a ViT-B architecture (with 12 layers) for both methods to ensure a fair comparison.

Two types of fine-tuning strategies are employed to gain insights into the pre-trained representations and design optimal fine-tuning approaches. Firstly, we use \textit{surgical} fine-tuning~\citep{lee2022surgical}, which fixes a fine-tuning budget (e.g., 3 layers) to update only specific layers, allowing us to identify the most effective layers to fine-tune. Secondly, we employ \textit{shallow} fine-tuning, which fine-tunes a shallower network consisting of only the first $N$ layers. This strategy updates the entire (shallow) network in contrast to the selective updates of \textit{surgical} fine-tuning, but is still more memory and compute efficient than end-to-end fine-tuning (see Figure~\ref{fig:pipeline} for a visual illustration).

Given a fixed fine-tuning budget, our \textit{surgical} fine-tuning results reveal that fine-tuning intermediate layers is more effective than the conventional wisdom of fine-tuning only the last few layers. Since different SSL methods learn different types of features, we find that fine-tuning the second quarter of the network (i.e., layers 4-6) is optimal for contrastive methods, while fine-tuning the third quarter of the network (i.e., layers 7-9) is optimal for restorative methods. These findings consistently hold true across a wide range of fine-tuning examples, ranging from 100 to 75K. Interestingly, fine-tuning only these three layers yields even better results than end-to-end fine-tuning, which fine-tunes the entire network. Intuitively, selectively optimizing a few layers while preserving well-learned representations in the remaining layers is advantageous, especially when working with relatively small fine-tuning datasets. Overall, however, we find that \textit{shallow} fine-tuning with layers 1-9 outperforms both \textit{surgical} and end-to-end fine-tuning, serving as the optimal fine-tuning strategy for both contrastive and restorative SSL across in-distribution as well as out-of-distribution transfer. Lastly, we show that our fine-tuning strategies not only enhance the performance of individual SSL models but also facilitate the complementary integration of multiple off-the-shelf SSL models, unlocking the potential for substantial improvements in self-supervised medical imaging.

To sum up, our main contributions include:
\begin{itemize}
    \item We present the first comprehensive study discovering effective \textit{fine-tuning} strategies for \textit{self-supervised} learning in \textit{medical imaging}.
    \item We establish strong contrastive and restorative SSL baselines for medical imaging, outperforming SOTA methods across four diverse tasks.
    \item Contrary to the conventional wisdom of fine-tuning only the last few layers, we show that fine-tuning intermediate layers yields far better results, with the exact layers depending on the pre-training algorithm (second/third quarter of network for contrastive/restorative SSL).
    \item Compared to the de-facto standard of end-to-end fine-tuning, we obtain improvements of as much as 5.48\% by fine-tuning a shallower network consisting of only the first three quarter layers of the pre-trained network.
    \item We propose a simple yet effective method to leverage the complementary strengths of multiple SSL models, resulting in gains of up to 3.57\% compared to using the best model alone.
\end{itemize}

\section{Related Work}
\label{sec:related_work}
Self-supervised learning enables models to learn generalizable representations from unlabeled data without the need for expensive manual annotations. Various forms of self-supervised learning have emerged in recent years, with contrastive and restorative methods being the most popular. Contrastive learning~\citep{hadsell2006dimensionality,chen2020simple,he2020momentum,chen2021exploring,chen2020big,zbontar2021barlow} learns meaningful representations by mapping the representations of similar images closer together whereas those of dissimilar images farther apart in the embedding space. On the other hand, restorative self-supervised learning~\citep{vincent2010stacked,pathak2016context,dosovitskiy2020vit,he2022masked}, also known as masked self-supervised learning, operates via reconstructing missing regions from a partially masked input.

Since collecting large labeled datasets is especially difficult in the context of medical imaging, self-supervised learning has proven to be extremely effective for learning representations from unlabeled medical data~\citep{azizi2021big,chaitanya2020contrastive,chen2019self,zhou2021models,tao2020revisiting,haghighi2021transferable,taher2022caid,zhou2021preservational,azizi2022robust,haghighi2022dira}. However, despite these advancements in self-supervised \textit{pre-training} algorithms, end-to-end fine-tuning still remains the dominant \textit{fine-tuning} strategy used in several SOTA medical imaging methods~\citep{taher2022caid,haghighi2022dira,azizi2022robust}. In this paper, we present the first comprehensive study designing optimal fine-tuning strategies for self-supervised medical imaging, demonstrating that our best fine-tuning strategies significantly outperform commonly used fine-tuning approaches. Moreover, using the insights from our fine-tuning analysis, we propose a simple yet effective method to directly leverage the complementary features of multiple self-supervised pre-trained models without any need for additional pre-training or delicate hyper-parameter tuning, as in previous methods~\citep{taher2022caid,haghighi2022dira}.

For a more detailed background of SSL and transfer learning, please refer to Appendix~\ref{sec:appendix_related_work}.

\section{Method}
\label{sec:method}
Section~\ref{sec:ssl} introduces the self-supervised methods, followed by detailed explanations of the pre-training and fine-tuning procedures in Sections~\ref{sec:pretraining} and~\ref{sec:finetuning} respectively. Lastly, Section~\ref{sec:finetuning_strategies} delves into the various fine-tuning strategies employed to effectively utilize the pre-trained features.
\subsection{Self-supervised Methods}
\label{sec:ssl}
In this work, we leverage MoCo-v3~\citep{chen2021empirical} and MAE~\citep{he2022masked} -- contrastive and restorative self-supervised methods built upon the vision transformer architecture~\citep{dosovitskiy2020vit} -- to develop strong baselines for self-supervised medical imaging. Building upon these baselines, we then design effective fine-tuning strategies, outperforming commonly used fine-tuning approaches by a significant margin. A detailed overview of both SSL methods is presented in Appendix~\ref{sec:mocov3_mae}.

\subsection{Pre-training}
\label{sec:pretraining}
We perform a comprehensive set of pre-training experiments where we pre-train both MoCo-v3 and MAE on ChestXray14~\citep{wang2017chestx} as well as the aggregated dataset (ChestXray14 + CheXpert~\citep{irvin2019chexpert}). Even though no labels are used during self-supervised pre-training, we only pre-train on the training splits of the respective datasets to avoid any potential data leakage from the validation and test sets. To ensure a fair comparison of pre-trained features, we employ a ViT-B/16 encoder (ViT-Base with a 16$\times$16 patch size) for both MoCo-v3 and MAE. Additionally, we apply the same augmentations as~\citet{hosseinzadeh2021systematic} for both methods. Specifically, we first crop and resize the images to 224$\times$224, followed by random horizontal flip (p$=$0.5) and random rotation (-7 to +7 degrees). We adopt the same settings specified in the official papers and train both the self-supervised models on 8 V100 GPUs. Lastly, we select the checkpoint with the lowest self-supervised loss within the last 5\% epochs for subsequent fine-tuning.
\subsection{Fine-tuning}
\label{sec:finetuning}
We evaluate the performance of the pre-trained models across a diverse set of medical imaging datasets encompassing both classification (ChestXray14, CheXpert) and segmentation (SIIM-ACR\footnote{https://www.kaggle.com/c/siim-acr-pneumothorax-segmentation/}, NIH Montgomery~\citep{jaeger2014two}) tasks. Similar to DiRA~\citep{haghighi2022dira}, we use the official data splits whenever available. Otherwise, we divide the datasets into 70/10/20 training/validation/test splits. More details about the datasets are presented in Appendix~\ref{sec:finetuning_datasets}.

For classification tasks, we use the pre-trained ViT-B encoder followed by a linear layer for fine-tuning. On the other hand, for segmentation tasks, we utilize a UNETR~\citep{hatamizadeh2022unetr} architecture, initializing the ViT-B encoder with the pre-trained weights while randomly initializing the decoder. AUC is used to evaluate classification tasks, whereas the Dice coefficient is used for segmentation tasks. We adopt the same fine-tuning settings as MAE~\citep{he2022masked} and fine-tune on a single V100 GPU.
\subsection{Fine-tuning Strategies}
\label{sec:finetuning_strategies}
We employ two types of fine-tuning strategies, namely \textit{surgical}~\citep{lee2022surgical} and \textit{shallow} fine-tuning, to gain insights into the pre-trained representations and design optimal fine-tuning approaches for self-supervised medical imaging analysis (see Figure~\ref{fig:pipeline} in the Appendix).

\textit{Surgical} fine-tuning uses a fixed fine-tuning budget (e.g., 3 layers) to update only a few specific layers while keeping the remaining layers frozen. By freezing some layers, we take advantage of pre-trained layers that already have effective representations for the downstream task. This approach is particularly beneficial when limited downstream data is available, as it avoids unnecessary updates to layers with almost optimal representations. We evaluate four different settings using the ViT-B model (consisting of 12 layers) -- fine-tuning layers 1-3, 4-6, 7-9, and 10-12. The latter setting (10-12) corresponds to the common practice of updating only the last few layers, allowing us to compare against this widely used fine-tuning technique. For each of the four settings, we perform extensive experiments with multiple pre-training and fine-tuning datasets as well as various fine-tuning dataset sizes ranging from 100 to 75K to understand the impact of various factors on the optimal layers to fine-tune. Each configuration is run at least three times, and we report the mean and standard deviation across all runs.

\textit{Shallow} fine-tuning, on the other hand, fine-tunes a shallower network consisting of only the first $N$ layers, but updates all layers within this shallower network (see Figure~\ref{fig:pipeline} in the Appendix). This strategy has the advantage that it updates the entire network in contrast to the selective updates of \textit{surgical} fine-tuning, while still being more memory and compute efficient than end-to-end fine-tuning. We experiment with four variations of shallow networks based on the ViT-B model, corresponding to layers 1-3, 1-6, 1-9, and 1-12 respectively. The last setting corresponds to the de-facto standard of end-to-end fine-tuning, allowing us to compare against this widely used fine-tuning technique. Similar to \textit{surgical} fine-tuning, each configuration is run at least three times, and we report the mean and standard deviation across all runs.

\section{Results}
\label{sec:experiments}
\begin{table}[t]\centering
\caption{Our strong SSL baselines for medical imaging surpass SOTA methods across 4 diverse tasks. Mean is reported across 3 runs for all experiments. For complete results (with std), please see Table~\ref{table:appendix_sota} in Appendix. ImageNet and CXray14 denote supervised pre-training via ImageNet and ChestXray14 respectively. CX14: ChestXray14, CXPT: CheXpert, SIIM: SIIM-ACR, Mont: NIH Montgomery.}
\label{table:sota}
\begin{tabular}{lllll}\toprule
\textbf{Method} &\textbf{CX14} &\textbf{CXPT} &\textbf{SIIM} &\textbf{Mont} \\\midrule
Random &80.31 &86.62 &67.54 &97.55 \\
ImageNet &81.70 &87.17 &67.93 &98.19 \\
CXray14 &NA &87.40 &68.92 &98.16 \\
\hdashline
CAiD &80.86 &87.44 &69.83 &98.19 \\
DiRA &81.12 &87.59 &69.87 &98.24 \\
\hdashline
MoCo-v3 & 81.35 & \textbf{87.77} & 77.19 & 98.22 \\
MAE &\textbf{82.68} &87.75&\textbf{79.25} &\textbf{98.28} \\
\bottomrule
\end{tabular}
\end{table}
\begin{figure*}[t]
\begin{minipage}{0.49\linewidth}
\centering
    \includegraphics[width=\textwidth]{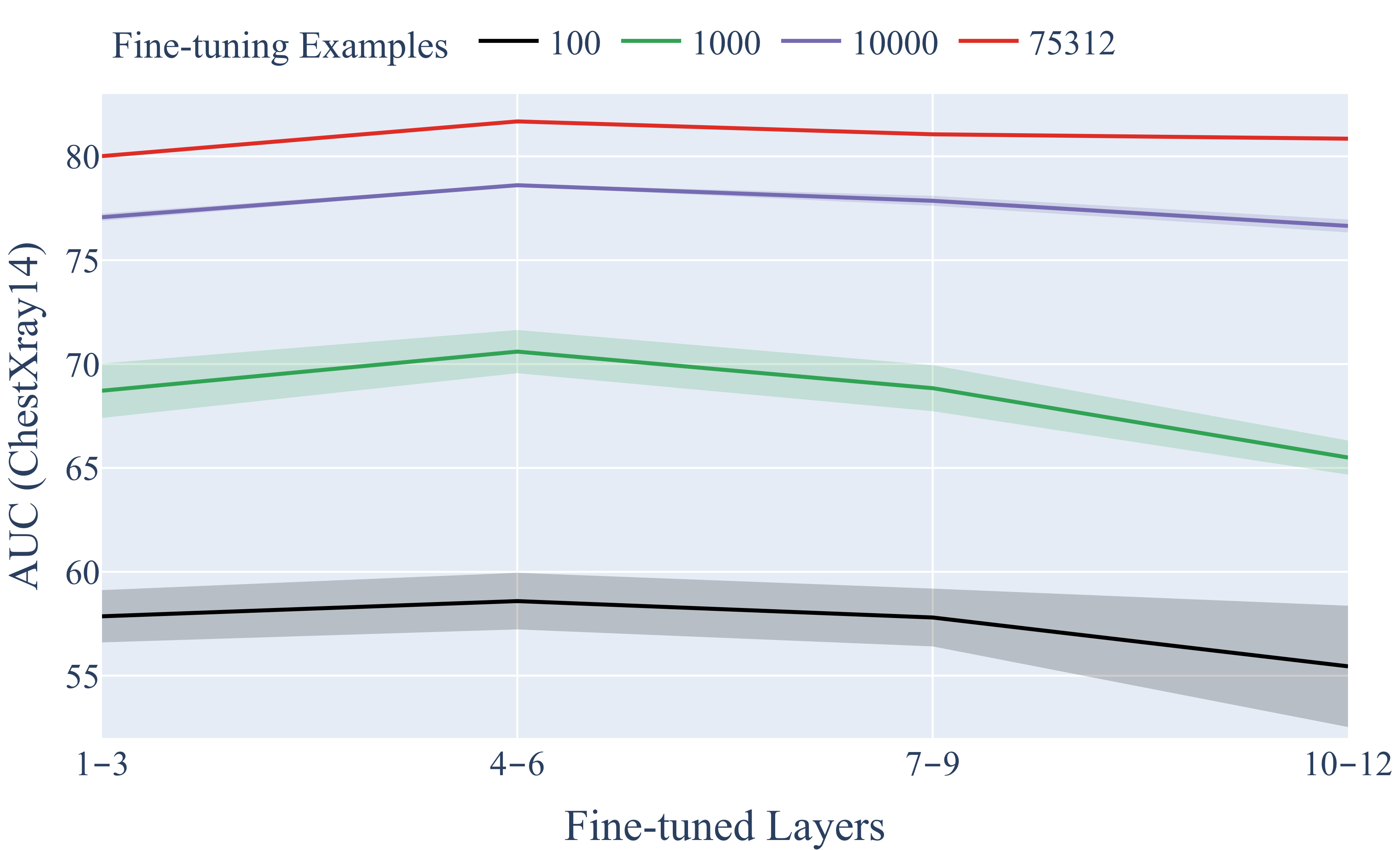}
\end{minipage}\hfill
\begin{minipage}{0.49\linewidth}
\centering
    \includegraphics[width=\textwidth]{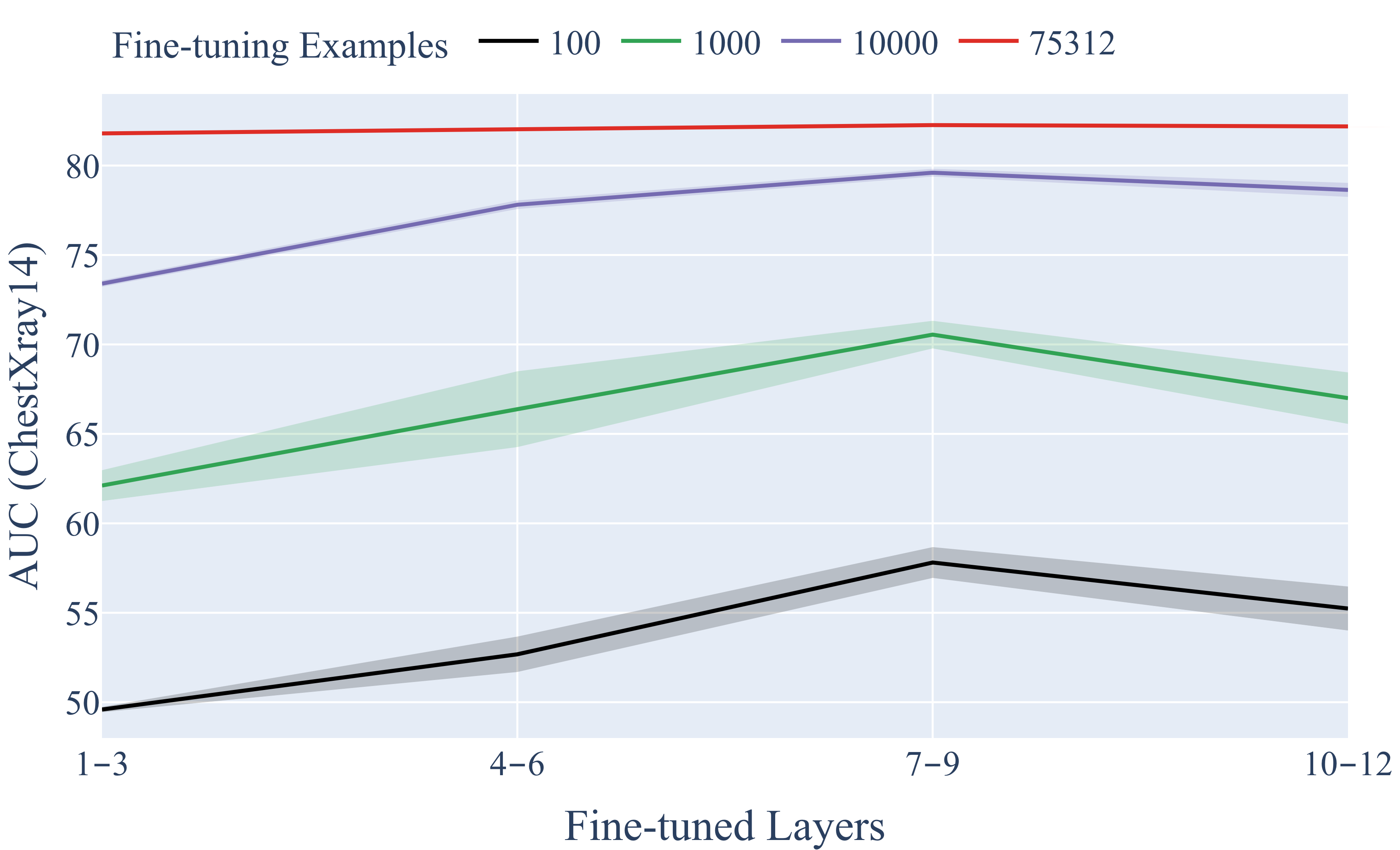}
\end{minipage}
\caption{Contrary to the conventional wisdom of fine-tuning the last few layers (10-12), we find that fine-tuning the intermediate layers is much more effective for utilizing self-supervised pre-trained features. \textbf{Left}: Fine-tuning the second quarter (i.e., 4-6) layers is optimal for contrastive (MoCo-v3) SSL. \textbf{Right}: Fine-tuning the third quarter (i.e., 7-9) layers is optimal for restorative (MAE) SSL. Mean and standard deviation are reported across 3 runs for all experiments.}
\label{fig:linecharts}
\end{figure*}
We conduct a comprehensive set of experiments to develop effective fine-tuning strategies for self-supervised medical imaging analysis. Firstly, in Section~\ref{sec:sota}, we establish strong contrastive and restorative self-supervised baselines for medical imaging, outperforming SOTA methods across four diverse downstream tasks. Building upon these strong baselines, Section~\ref{sec:optimal_ft} designs optimal fine-tuning strategies that significantly outperform commonly used fine-tuning approaches across various pre-training and fine-tuning datasets. Next, using the insights from this analysis, Section~\ref{sec:complementarity} proposes a simple yet effective method to leverage the complementary strengths of multiple SSL models without requiring any additional pre-training. Lastly, Appendix~\ref{sec:ablations} presents various ablation studies to further validate our findings.
\subsection{Contrastive and Restorative Medical Imaging SSL}
\label{sec:sota}
In this section, we set up strong contrastive and restorative self-supervised baselines for medical imaging based on the robust self-supervised learning methods MoCo-v3~\citep{chen2021empirical} and MAE~\citep{he2022masked}. Table~\ref{table:sota} presents comprehensive results illustrating the efficacy of these methods on the four diverse downstream tasks introduced in Section~\ref{sec:finetuning}.

To evaluate the effectiveness of our baselines, we compare our results against two SOTA SSL methods for medical imaging: CAiD~\citep{taher2022caid} and DiRA~\citep{haghighi2022dira}. Both these SOTA methods utilize multiple types of self-supervised learning, with CAiD~\citep{taher2022caid} leveraging contrastive and restorative SSL, while DiRA~\citep{haghighi2022dira} benefiting from contrastive, restorative, as well as adversarial learning. Our comprehensive results across four fine-tuning datasets demonstrate that our baselines surpass both methods, even without employing any additional techniques such as joint training with multiple types of SSL. It is worth noting that the reported results of CAiD~\citep{taher2022caid} and DiRA~\citep{haghighi2022dira} in Table~\ref{table:sota} actually represent the best performance among three self-supervised methods [MoCo-v2~\citep{chen2020improved}, Barlow Twins~\citep{zbontar2021barlow}, and SimSiam~\citep{chen2021exploring}]. Nevertheless, we consistently outperform both CAiD~\citep{taher2022caid} and DiRA~\citep{haghighi2022dira} on all four downstream datasets, illustrating the effectiveness of our strong self-supervised baselines for medical imaging.

Although our baselines achieve excellent performance across a diverse range of tasks, it remains unclear whether the default end-to-end fine-tuning is the best approach to leverage pre-trained features. This is especially true since different SSL methods capture distinct features, potentially requiring unique fine-tuning strategies for optimal utilization of pre-trained knowledge. We explore this further in the next section.
\subsection{Optimal Fine-tuning Strategies}
\label{sec:optimal_ft}
\begin{figure*}[t]
\begin{minipage}{.495\linewidth}
\centering
    \includegraphics[width=\textwidth]{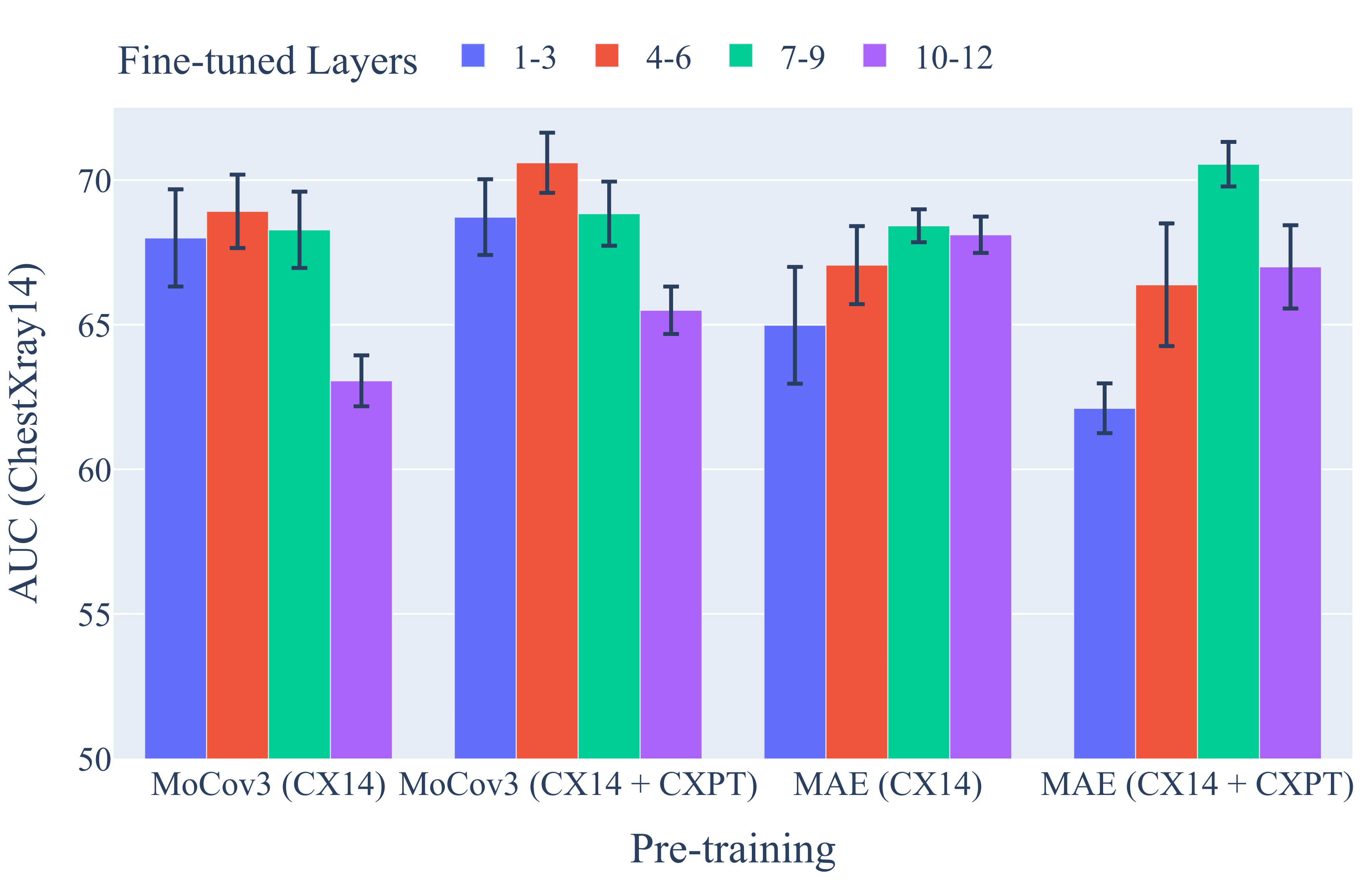}
\end{minipage}\hfill
\begin{minipage}{.495\linewidth}
\centering
    \includegraphics[width=\textwidth]{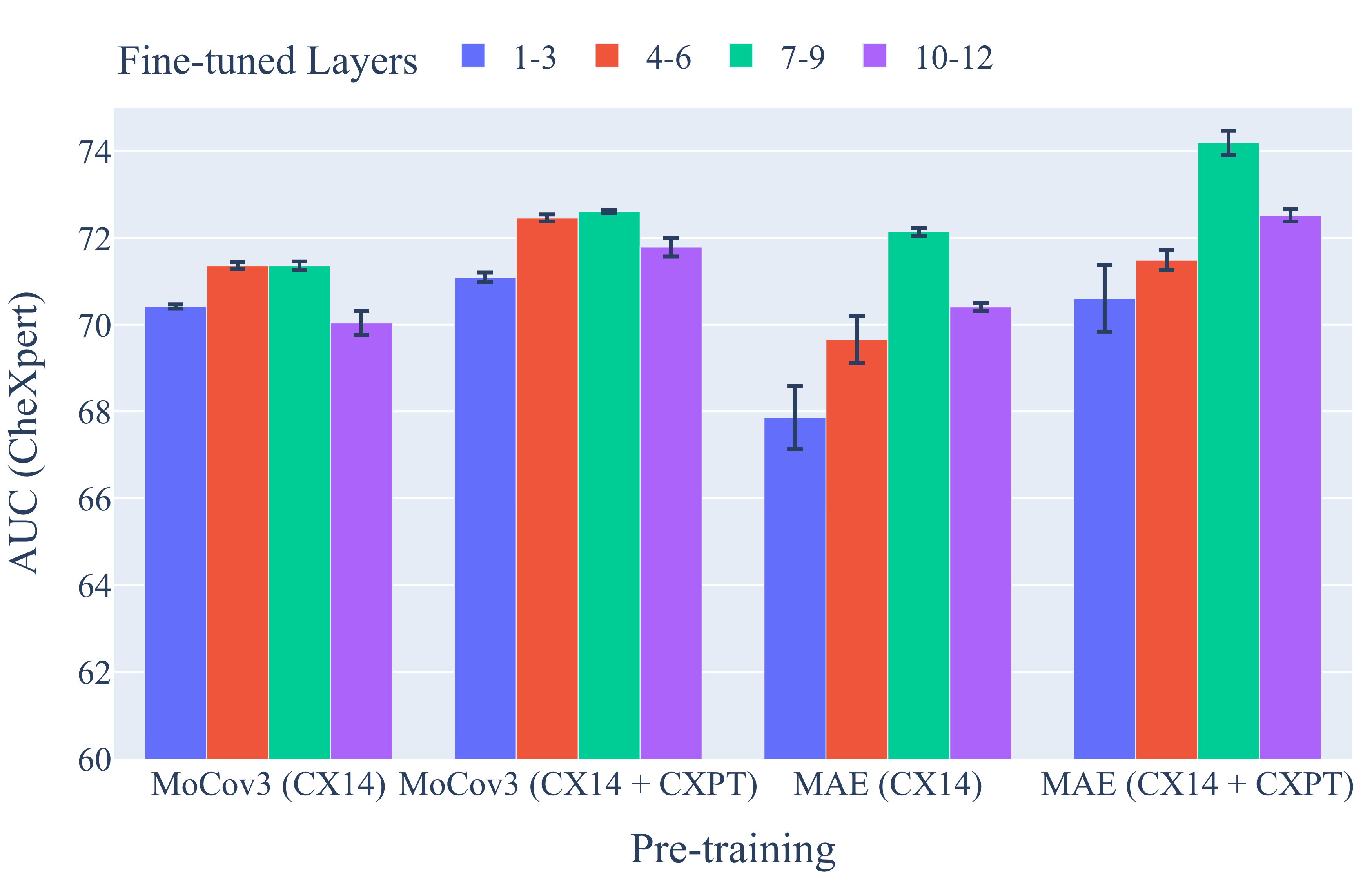}
\end{minipage}
\caption{Effect of pre-training and fine-tuning datasets on the optimal \textit{surgical} fine-tuning strategy. Fine-tuning second quarter (4-6) layers is optimal for contrastive (MoCo-v3) SSL whereas fine-tuning third quarter (7-9) layers is optimal for restorative (MAE) SSL, across all pre-training and fine-tuning datasets. Mean and standard deviation are reported across 3 runs for all experiments. \textbf{Left}: Fine-tuned on ChestXray14. \textbf{Right}: Fine-tuned on CheXpert. CX14: ChestXray14, CXPT: CheXpert.}
\label{fig:barcharts_surgical}
\end{figure*}
In this section, we discover the optimal fine-tuning strategies for contrastive and restorative self-supervised learning using the fine-tuning techniques introduced in Section~\ref{sec:finetuning_strategies}.

\subsubsection{Best Layers to Fine-tune}
Firstly, since contrastive and restorative methods optimize for different pre-training objectives, potentially learning distinct features, we seek to answer the following questions: Given a fixed fine-tuning budget (e.g., 3 layers), which layers are best to fine-tune for \textit{contrastive} and \textit{restorative} methods? How does the optimal setup compare to the commonly used technique of only fine-tuning the last few layers (e.g., layers 10--12)? What is the effect of the fine-tuning dataset size on the optimal layers to fine-tune? To answer these questions, we conduct pre-training using MoCo-v3 and MAE on the aggregated (ChestXray14 + CheXpert) dataset and perform extensive \textit{surgical} fine-tuning experiments on ChestXray14, utilizing fine-tuning datasets of various sizes: 100, 1000, 10000, and 75312 (full dataset).

Figure~\ref{fig:linecharts} illustrates the efficacy of fine-tuning layers at various depths within both the MoCo-v3 and MAE pre-trained networks. Given a fixed fine-tuning budget (3 layers), we find that fine-tuning intermediate layers yields significantly better results compared to the common practice of fine-tuning the last few layers. Specifically, we discover that fine-tuning the second quarter of the network (i.e., layers 4-6) is optimal for contrastive pre-training methods such as MoCo-v3 whereas fine-tuning the third quarter of the network (i.e., layers 7-9) is optimal for restorative pre-training methods such as MAE. These findings consistently hold true across a wide range of fine-tuning examples ranging from 100 to 75K, with the difference between fine-tuning intermediate layers and last layers becoming especially pronounced when fine-tuning on smaller datasets. Remarkably, fine-tuning only these three intermediate layers can achieve even better results than end-to-end fine-tuning, which fine-tunes the entire network (Section~\ref{sec:comparison_e2e_ft}). Intuitively, selectively optimizing a few layers avoids unnecessary updates to the remaining layers, thereby preserving well-learned representations, which is especially beneficial when working with relatively small fine-tuning datasets.

\begin{figure*}[t]
\begin{minipage}{.495\linewidth}
\centering
    \includegraphics[width=\textwidth]{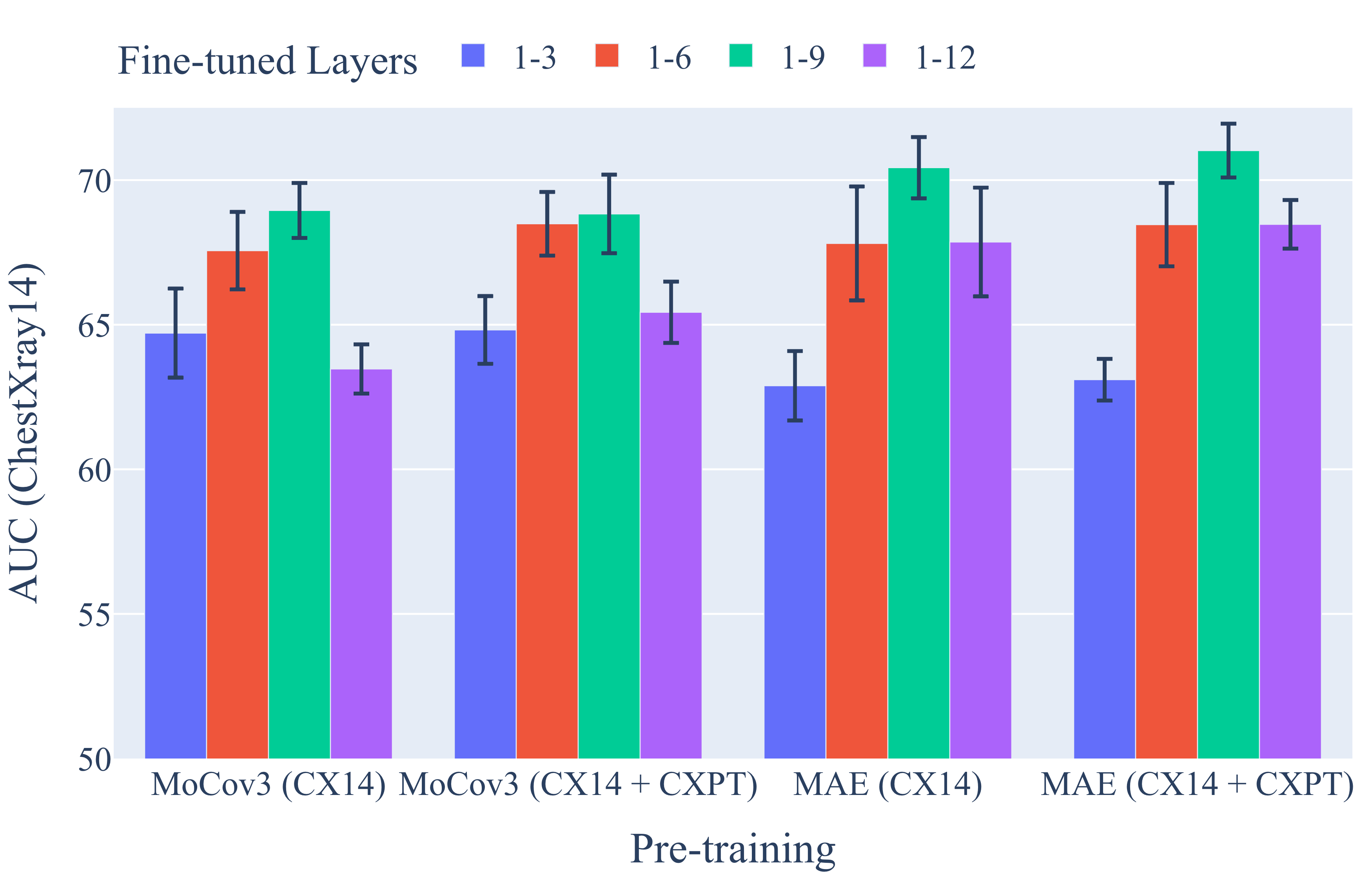}
\end{minipage}\hfill
\begin{minipage}{.495\linewidth}
\centering
    \includegraphics[width=\textwidth]{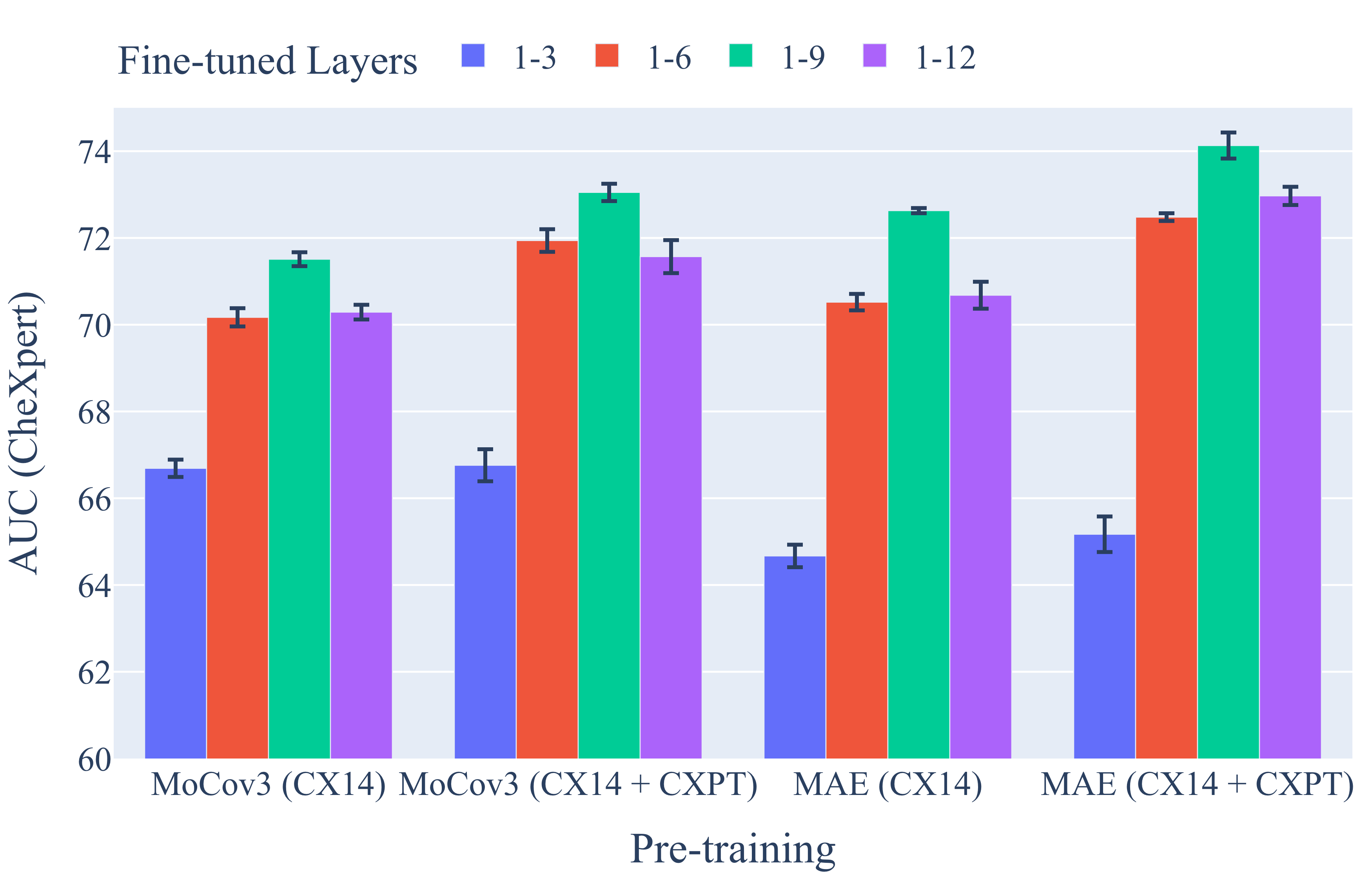}
\end{minipage}
\caption{Effect of pre-training and fine-tuning datasets on the optimal \textit{shallow} fine-tuning strategy. Fine-tuning the shallow network consisting of layers 1-9 is optimal for both contrastive (MoCo-v3) and restorative (MAE) SSL, across all pre-training and fine-tuning datasets. Moreover, it consistently outperforms the de-facto standard of end-to-end fine-tuning across all settings (green vs purple bars). Mean and standard deviation are reported across 3 runs for all experiments. \textbf{Left}: Fine-tuned on ChestXray14. \textbf{Right}: Fine-tuned on CheXpert. CX14: ChestXray14, CXPT: CheXpert.}
\label{fig:barcharts_shallow}
\end{figure*}
\begin{table*}[t]\centering
\caption{Optimal \textit{surgical} and \textit{shallow} fine-tuning strategies compared against the default end-to-end (E2E) fine-tuning setup. Contrastive (MoCo) and restorative (MAE) SSL methods are pre-trained on ChestXray14, and subsequently fine-tuned on ChestXray14 (in-distribution transfer) and CheXpert (out-of-distribution transfer). \textit{Shallow} fine-tuning with layers 1-9 is the optimal fine-tuning strategy across both contrastive and restorative methods and across in-distribution as well as out-of-distribution transfer. Mean is reported across 3 runs for all experiments. Standard deviation is omitted for clarity. FT: fine-tuning.
}\label{tab:comparison_e2e}
\footnotesize
\begin{tabular}{@{}lcccccccccccc@{}}\toprule
&\multicolumn{6}{c}{\textbf{MoCo (ChestXray14)}} &\multicolumn{6}{c}{\textbf{MAE (ChestXray14)}} \\\cmidrule(lr){2-7} \cmidrule(lr){8-13}
\multirow{2}{*}{\textbf{FT Dataset}} &\multicolumn{2}{c}{E2E FT} &\multicolumn{2}{c}{Surgical FT} &\multicolumn{2}{c}{Shallow FT} &\multicolumn{2}{c}{E2E FT} &\multicolumn{2}{c}{Surgical FT} &\multicolumn{2}{c}{Shallow FT} \\\cmidrule{2-3} \cmidrule(lr){4-5} \cmidrule(lr){6-7} \cmidrule{8-9} \cmidrule(lr){10-11} \cmidrule(lr){12-13}
&Layers &AUC &Layers &AUC &Layers &AUC &Layers &AUC &Layers &AUC &Layers &AUC \\\midrule
\textbf{ChestXray14} &1-12 &63.47 &4-6 &68.92 &1-9 &\textbf{68.95} &1-12 &67.86 &7-9 &68.42 &1-9 &\textbf{70.43} \\
\textbf{CheXpert} &1-12 &70.29 &4-6 &71.36 &1-9 &\textbf{71.51} &1-12 &70.68 &7-9 &72.14 &1-9 &\textbf{72.63} \\
\bottomrule
\end{tabular}
\end{table*}
\subsubsection{Effect of Pre-training and Fine-tuning Datasets}
Here, we investigate the impact of different pre-training and fine-tuning datasets on the optimal fine-tuning configuration for both \textit{surgical} and \textit{shallow} fine-tuning. To this end, we pre-train both MoCo-v3 and MAE using the pre-training datasets ChestXray14 and (ChestXray14 + CheXpert) and evaluate using the fine-tuning datasets ChestXray14 and CheXpert, employing 1000 fine-tuning examples in all experiments. Figures~\ref{fig:barcharts_surgical} and~\ref{fig:barcharts_shallow} illustrate the \textit{surgical} and \textit{shallow} fine-tuning results respectively.

From the \textit{surgical} fine-tuning results in Figure~\ref{fig:barcharts_surgical}, we observe that increasing the pre-training dataset size from ChestXray14 to (ChestXray14 + CheXpert) improves the downstream performance for both contrastive and restorative pre-trained models across both downstream tasks. While the choice of pre-training dataset affects the quantitative downstream performance, it does not impact the optimal fine-tuning strategy. Specifically, for contrastive methods (MoCo-v3), we find that fine-tuning the second quarter (i.e., 4-6) layers is significantly better than fine-tuning the last few layers. On the other hand, for restorative methods (MAE), we discover that fine-tuning the third quarter (i.e., 7-9) layers offers the best downstream performance across various pre-training and fine-tuning datasets. Overall, given a fixed fine-tuning budget, we conclude that fine-tuning intermediate layers (e.g., second/third quarter layers) consistently outperforms the commonly used technique of fine-tuning the last few layers for both contrastive and restorative methods, regardless of the pre-training and fine-tuning datasets.

Based on the \textit{shallow} fine-tuning results in Figure~\ref{fig:barcharts_shallow}, we discover that fine-tuning the shallow network consisting of layers 1-9 (i.e., excluding the last 3 layers) is the optimal configuration for both contrastive and restorative methods across all pre-training and fine-tuning datasets. Moreover, it is significantly more effective than the de-facto standard of end-to-end fine-tuning. Interestingly, end-to-end fine-tuning with 12 layers (purple bar) is sometimes even worse than fine-tuning a quarter-sized network consisting of only the first 3 layers (blue bar). This can be seen from the leftmost subgroup in Figure~\ref{fig:barcharts_shallow} (left) (i.e., MoCo-v3 (CX14)), which illustrates that the last quarter layers are particularly challenging to adapt to the downstream task, and their inclusion leads to a significant drop in overall network performance. This could be attributed to the fact that the last few layers specialize in solving the self-supervised pretext task and hence fail to capture general representations that are useful for various downstream tasks. In conclusion, fine-tuning a shallower network comprising layers 1-9 significantly outperforms the commonly used technique of end-to-end fine-tuning for both contrastive and restorative self-supervised learning, regardless of the pre-training and fine-tuning datasets. We present a detailed comparison against the default end-to-end fine-tuning strategy in the next section.
\subsubsection{Comparison with End-to-End Fine-tuning}
\label{sec:comparison_e2e_ft}
\begin{table*}[t]\centering
\caption{Leveraging the complementary features of contrastive (MoCo) and restorative (MAE) self-supervised pre-trained models. Our methods yield consistent improvements and significantly outperform the naive complementary strategy (MAE12 + MoCo12) across multiple downstream tasks and fine-tuning dataset sizes. Mean and standard deviation are reported across 3 runs for all experiments.}\label{table:complementarity}
\footnotesize
\begin{tabular}{lllllll}\toprule
\multirow{3}{*}{\textbf{Method}} & \multirow{3}{*}{\textbf{FT Strategy}} & \multirow{3}{*}{\textbf{FT Layers}} &\multicolumn{4}{c}{\textbf{Fine-tuning Examples}} \\
&&&\multicolumn{2}{c}{100} &\multicolumn{2}{c}{1000} \\\cmidrule(lr){4-5} \cmidrule(lr){6-7}
&&&ChestXray14 &CheXpert &ChestXray14 &CheXpert \\\midrule
\textbf{MoCo} & End-to-end & 1-12 &54.16±2.04 & 63.33±0.72 & 65.19±0.97 & 71.59±0.43 \\
\textbf{MAE} & End-to-end & 1-12 &54.91±0.68 & 63.56±1.20 & 68.84±0.58 & 72.84±0.12 \\
\hdashline
\textbf{MAE12 + MoCo12} & End-to-end & 1-12/1-12 &56.61±0.43 &65.14±1.52 &69.24±1.41 &73.16±0.46 \\
\hdashline
\textbf{MAE9 + MoCo9} & Shallow & 1-9/1-9 &\textbf{58.48±1.69} &\textbf{66.07±0.87} &71.43±0.39 &73.83±0.29 \\
\textbf{MAE9 + MoCo6} & Surgical  & 7-9/4-6 &58.01±1.83 &65.41±1.04 &\textbf{71.71±0.67} &\textbf{74.25±0.06} \\
\bottomrule
\end{tabular}
\end{table*}
In this section, we compare the optimal \textit{surgical} and \textit{shallow} fine-tuning approaches against the default end-to-end (E2E) fine-tuning setup in both the in-distribution and out-of-distribution transfer settings. Table~\ref{tab:comparison_e2e} summarizes the main findings. \textit{Surgical} fine-tuning consistently outperforms E2E fine-tuning for both pre-training methods and in both evaluation settings, with fine-tuning the second quarter of the network being optimal for contrastive (MoCo) methods whereas fine-tuning the third quarter of the network being optimal for restorative (MAE) methods. Although \textit{surgical} fine-tuning already yields substantial improvements over E2E fine-tuning, \textit{shallow} fine-tuning with layers 1-9 is the best fine-tuning strategy across both contrastive and restorative self-supervised methods and in both in-distribution and out-of-distribution transfer settings. Compared to the commonly used E2E fine-tuning strategy, \textit{shallow} fine-tuning yields significant gains in both evaluation settings, with gains of as much as 5.48\% in the in-distribution setting. These substantial improvements highlight the effectiveness of our fine-tuning strategies in maximizing the utility of pre-trained features for self-supervised medical imaging analysis.
\subsection{Complementarity}
\label{sec:complementarity}
Here, we propose a simple yet effective method to leverage the complementary strengths of multiple SSL models by using the insights from the optimal fine-tuning strategies developed in the previous section.

Firstly, we establish a simple baseline for leveraging the complementary features of the MAE and MoCo-v3 pre-trained models. This involves adding a linear classifier on top of the concatenated features from the last layers (i.e., layers 12) of the ViT-B trained encoders, and training the entire network end-to-end. We refer to this baseline as MAE12 + MoCo12 in Table~\ref{table:complementarity}. Although this strategy yields better results than either of the pre-trained models alone, it fails to fully utilize the complementary features of the two pre-trained models.

Inspired by the insight that fine-tuning intermediate layers leads to better feature utilization in both contrastive and restorative SSL methods, we devise two strategies to optimally leverage the complementary features of the MAE and MoCo pre-trained models. Firstly, instead of using the entire pre-trained MAE and MoCo encoders, we use shallower networks comprised of layers 1-9 and fine-tune these shallower networks along with a linear classifier on top of the extracted features (MAE9 + MoCo9 (Shallow)). Secondly, we employ \textit{surgical} fine-tuning and only fine-tune the third quarter layers for MAE and second quarter layers for MoCo (MAE9 + MoCo6 (Surgical)). As illustrated in Table~\ref{table:complementarity}, our proposed strategies effectively utilize the complementary strengths of both contrastive and restorative pre-trained models, leading to improvements of as much as 3.57\% compared to using the best model (MAE) alone.

Hence, our fine-tuning techniques not only enhance the performance of individual SSL models, but also offer a simple approach to directly leverage the complementary features of multiple pre-trained SSL models without any need for joint pre-training or delicate hyper-parameter tuning, as in previous methods~\citep{taher2022caid,haghighi2022dira}.

Appendix~\ref{sec:ablations} presents ablation studies on model convergence, fine-tuning normalization and nearest neighbor analysis to further elucidate our findings.
\section{Conclusion}
\label{sec:conclusion}
In this paper, we presented the first comprehensive study discovering effective \textit{fine-tuning} strategies for \textit{self-supervised} learning in \textit{medical imaging}. We established strong contrastive and restorative SSL baselines that outperform SOTA methods across four diverse downstream tasks. Building on these, we then devised effective fine-tuning strategies that maximize the utility of pre-trained features and yield significant benefits compared to commonly used fine-tuning approaches. Moreover, using these insights, we proposed a simple yet effective method to leverage the complementary strengths of multiple SSL models. Given the rapid advances in SSL, our fine-tuning strategies not only have the potential to significantly enhance the performance of individual SSL models, but also enable effective utilization of the complementary strengths offered by multiple SSL models, leading to significant improvements in self-supervised medical imaging analysis.
\bibliography{jmlr-sample}
\clearpage
\appendix

\section{Ablations}
\label{sec:ablations}
\begin{table}[b]\centering
\caption{Convergence analysis of MoCo-v3 and MAE pre-trained models via \textit{surgical} fine-tuning (FT).}\label{tab:convergence_surgical}
\resizebox{\columnwidth}{!}{%
\begin{tabular}{@{}cccc@{}}\toprule
\multirow{2}{*}{\textbf{FT Layers}} &\multirow{2}{*}{\textbf{FT Params (M)}} &\multicolumn{2}{c}{\textbf{Convergence Epochs}} \\\cmidrule{3-4}
&&MoCo-v3 &MAE \\\midrule
1-3 & 22.0 &300 &957 \\
4-6 & 22.0 &133 &690 \\
7-9 & 22.0 &63 &103 \\
10-12 & 22.0 &67 &47 \\
\bottomrule
\end{tabular}
}
\end{table}
\begin{table}[t]\centering
\caption{Convergence analysis of MoCo-v3 and MAE pre-trained models via \textit{shallow} fine-tuning (FT).}\label{tab:convergence_shallow}
\resizebox{\columnwidth}{!}{%
\begin{tabular}{@{}cccc@{}}\toprule
\multirow{2}{*}{\textbf{FT Layers}} &\multirow{2}{*}{\textbf{FT Params (M)}} &\multicolumn{2}{c}{\textbf{Convergence Epochs}} \\\cmidrule{3-4}
&&MoCo-v3 &MAE \\\midrule
1-3 & 22.0 &417 &920 \\
1-6 & 43.3 &113 &303 \\
1-9 & 64.6 &57 &67 \\
1-12 & 85.8 &50 &63 \\
\bottomrule
\end{tabular}
}
\end{table}
\begin{table}[b]\centering
\caption{Normalization statistics of pre-training dataset versus fine-tuning dataset during fine-tuning (FT). CX14: ChestXray14, CXPT: CheXpert.}\label{tab:ft_normalization}
\resizebox{\columnwidth}{!}{%
\begin{tabular}{@{}llll@{}}\toprule
\textbf{Pre-train} &\textbf{FT Normalization} &\textbf{CX14} &\textbf{CXPT} \\\midrule
MAE &Pre-train Dataset &82.80 &87.70 \\
MAE &Fine-tune Dataset &82.79 &88.07 \\
\bottomrule
\end{tabular}
}
\end{table}
\subsection{Convergence Analysis}
\begin{figure*}[t]
  \centering
  \begin{tabular}{c}
      \includegraphics[width=\textwidth]{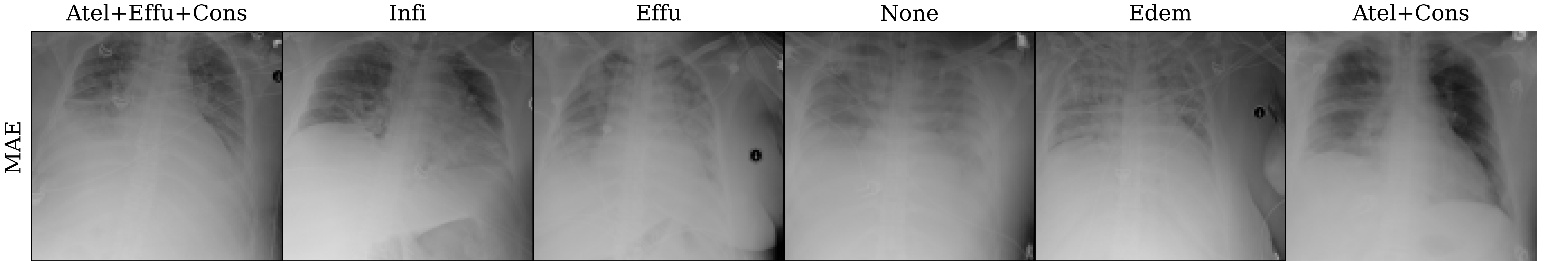} \\
      \includegraphics[width=\textwidth]{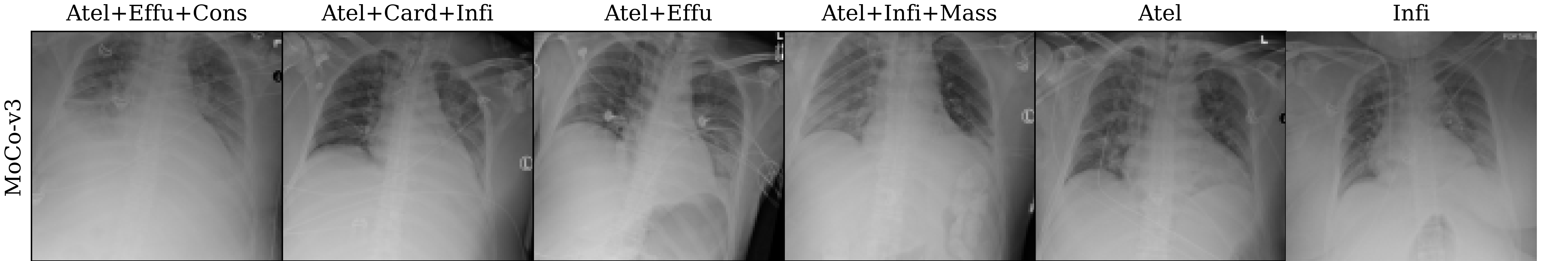} \\
      \includegraphics[width=\textwidth]{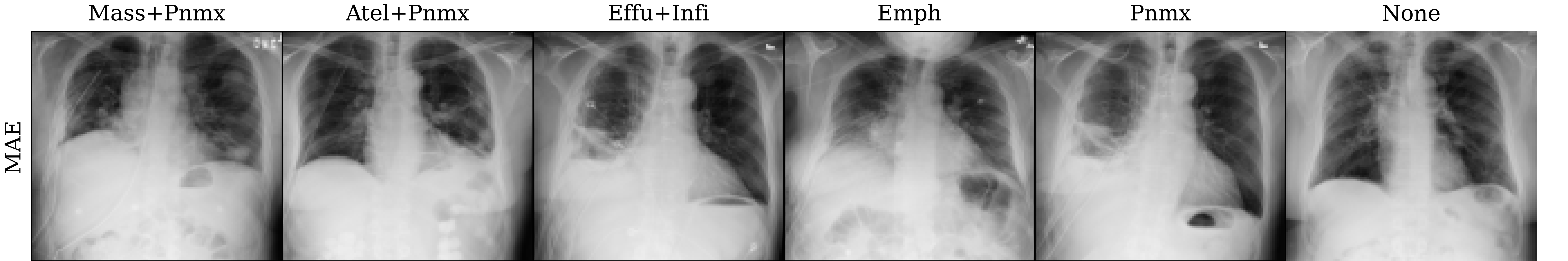} \\
      \includegraphics[width=\textwidth]{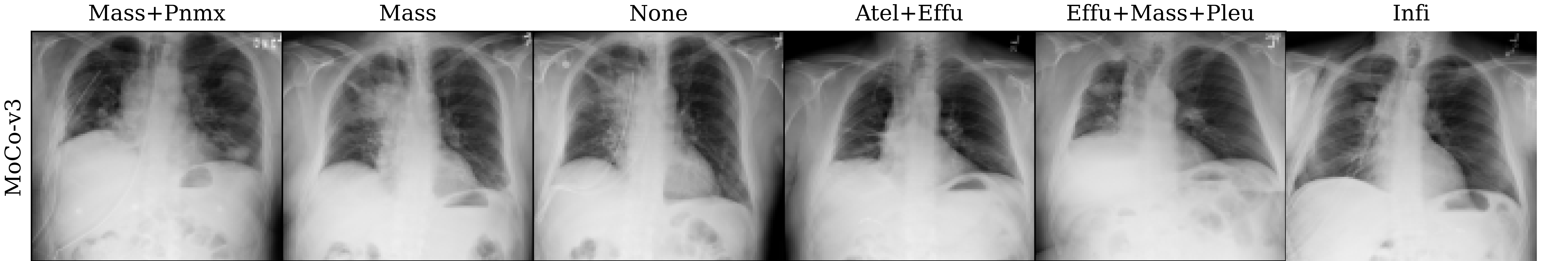} \\
  \end{tabular}
  \caption{Nearest neighbor visualizations of MAE (top) and MoCo-v3 (bottom). Leftmost image is the query image followed by its 5 nearest neighbors in the embedding space. Atel: Atelectasis, Card: Cardiomegaly, Cons: Consolidation, Edem: Edema, Effu: Effusion, Emph: Emphysema, Fibr: Fibrosis, Hern: Hernia, Infi: Infiltration, Mass: Mass, Nodu: Nodule, Pleu: Pleural Thickening, Pnma: Pneumonia, Pnmx: Pneumothorax, None: No disease.}
  \label{fig:nn_vis}
\end{figure*}
In this section, we investigate the convergence behavior of different fine-tuning strategies. Table~\ref{tab:convergence_surgical} and Table~\ref{tab:convergence_shallow} present the results for \textit{surgical} and \textit{shallow} fine-tuning respectively, with all experiments conducted on the ChestXray14 dataset. For \textit{surgical} fine-tuning (Table~\ref{tab:convergence_surgical}), we observe that fine-tuning the later layers significantly accelerates convergence. We hypothesize that fine-tuning only the earlier layers is slower because they additionally need to adjust their features so that they align well with the frozen features of the subsequent layers. Similarly, for \textit{shallow} fine-tuning (Table~\ref{tab:convergence_shallow}), we find that fine-tuning deeper networks leads to faster convergence since the entire network can adapt to the downstream task, with each layer specializing for a specific set of features. Moreover, we note that MoCo-v3 demonstrates faster convergence compared to MAE across both \textit{surgical} and \textit{shallow} fine-tuning.
\subsection{Fine-tuning Normalization}
Next, we conduct an ablation study to evaluate the effectiveness of using the normalization statistics of the pre-training dataset versus the fine-tuning dataset during the fine-tuning process. The results in Table~\ref{tab:ft_normalization} reveal that using the fine-tuning dataset normalization is at par or slightly better than utilizing the pre-training dataset normalization. Hence, we use this as the default setting for all the experiments in this paper.
\subsection{Nearest Neighbor Analysis}
Lastly, we visualize the nearest neighbors in the embedding space of MoCo-v3 and MAE pre-trained models to yield insights into the pre-trained representations. To this end, we extract the last layer representations (after the layer norm) for both the train and test sets of ChestXray14. For each image in the test set, we compute the 5 nearest neighbors from the train set. The results are visualized in Figure~\ref{fig:nn_vis}, which reveals that the nearest neighbors of both MoCo-v3 and MAE largely preserve the overall shape of the query image. Additionally, we also perform a quantitative analysis by assigning the label of the nearest neighbor as the prediction for the query image. Under this nearest neighbor classification setting, MoCo-v3 has a slightly higher mAUC than MAE (52.7 vs 52.3), indicating that its pre-trained features are more linearly separable. This finding is in line with the observations made by~\citet{he2022masked}, who observed that the MoCo-v3 features are more linearly separable than MAE features. However, it is worth noting that despite MAE features being less linearly separable than MoCo-v3 features, the former captures more powerful non-linear features. This translates to stronger downstream models when the networks are fine-tuned instead of being used as fixed feature extractors, as evidenced by the superior fine-tuning results presented in Section~\ref{sec:sota}.
\section{Implementation Details}
\begin{figure*}[t]
\centering
\includegraphics[width=0.89\textwidth]{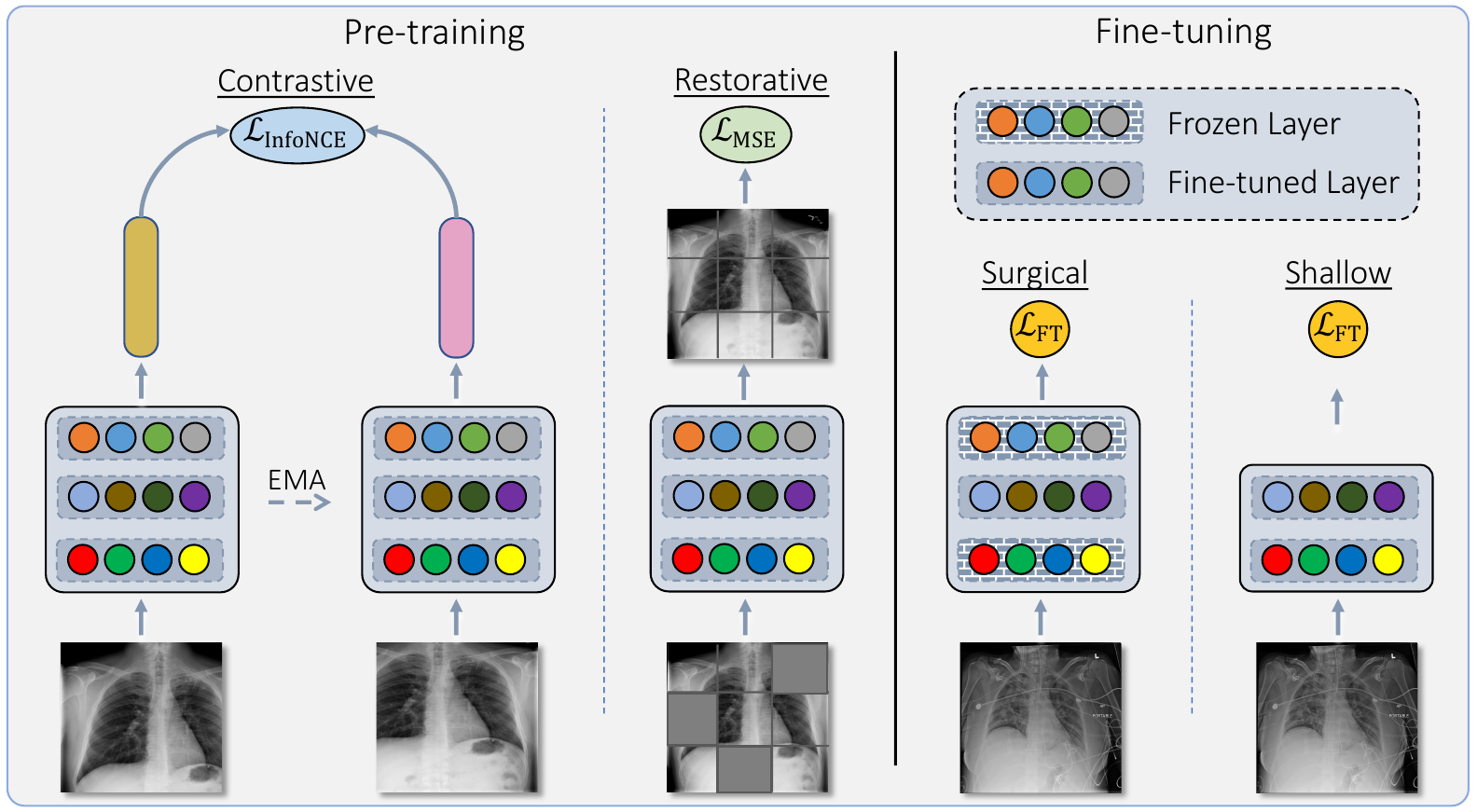}
\caption{We design optimal fine-tuning strategies for contrastive and restorative self-supervised medical imaging analysis, utilizing two distinct types of fine-tuning techniques. \textit{Surgical} fine-tuning updates only a few specific layers within the pre-trained network. On the other hand, \textit{shallow} fine-tuning employs a shallower network than the pre-trained network, but updates all layers within this shallow network. Our comprehensive evaluations demonstrate that our fine-tuning strategies significantly outperform commonly used fine-tuning techniques (Section~\ref{sec:optimal_ft}).}
\label{fig:pipeline}
\end{figure*}

Figure~\ref{fig:pipeline} provides a visual illustration of the pre-training methods and fine-tuning techniques employed in this study.
\begin{table*}[t]\centering
\caption{Our strong self-supervised baselines for medical imaging surpass SOTA methods across 4 diverse downstream tasks, even without employing any additional techniques such as joint training with multiple types of SSL. Mean and standard deviation are reported across 3 runs for all experiments. ImageNet and CXray14 denote supervised pre-training via ImageNet and ChestXray14 respectively.}
\label{table:appendix_sota}
\begin{tabular}{lllll}\toprule
\textbf{Method} &\textbf{ChestXray14} &\textbf{CheXpert} &\textbf{SIIM-ACR} &\textbf{Montgomery} \\\midrule
Random &80.31$\pm$0.10 &86.62$\pm$0.15 &67.54$\pm$0.60 &97.55$\pm$0.36 \\
ImageNet &81.70$\pm$0.15 &87.17$\pm$0.22 &67.93$\pm$1.45 &98.19$\pm$0.13 \\
CXray14 &NA &87.40$\pm$0.26 &68.92$\pm$0.98 &98.16$\pm$0.05 \\
\hdashline
CAiD &80.86$\pm$0.16 &87.44$\pm$0.33 &69.83$\pm$0.29 &98.19$\pm$0.08 \\
DiRA &81.12$\pm$0.17 &87.59$\pm$0.28 &69.87$\pm$0.68 &98.24$\pm$0.09 \\
\hdashline
MoCo-v3 & 81.35±0.09 & \textbf{87.77±0.49} & 77.19±0.73 & 98.22±0.05 \\
MAE &\textbf{82.68$\pm$0.07} &87.75$\pm$0.28 &\textbf{79.25$\pm$0.45} &\textbf{98.28$\pm$0.02} \\
\bottomrule
\end{tabular}
\end{table*}
\subsection{Pre-training Methods}
\label{sec:mocov3_mae}
MoCo-v3~\citep{chen2021empirical} builds upon the MoCo~\citep{he2020momentum,chen2020improved} framework, which we briefly review next. Given positive and negative pairs, MoCo-v3~\citep{chen2021empirical} encourages positive pairs to have similar representations whereas negative pairs to have dissimilar representations in the embedding space. This is accomplished via the contrastive InfoNCE loss~\citep{oord2018representation}, formulated as:
\begin{equation}
\mathcal{L}_\texttt{InfoNCE}=-\log \frac{\exp \left(q \cdot k^{+} / \tau\right)}{\exp \left(q \cdot k^{+} / \tau\right)+\sum_{i=1}^{N} \exp \left(q \cdot k_{i}^{-} / \tau\right)}
\end{equation}
where $q$ is the query image representation, $k^{+}$ is the positive image representation, \{$k_{1}^{-}, \ldots, k_{N}^{-}$\} is the set of representations of the $N$ negative images, and $\tau$ is a temperature hyper-parameter. Augmented versions of the same image form a positive pair whereas augmentations of different images form a negative pair. The query image representation $q$ is generated using an encoder $f$ while the positive and negative image representations ($k^{+}$ and $k_{i}^{-}$) are obtained using another encoder $g$. $f$ consists of a ViT~\citep{dosovitskiy2020vit} backbone followed by projection and prediction heads, whereas $g$ consists of only the ViT~\citep{dosovitskiy2020vit} backbone and the projection head. The parameters of $f$ are updated via gradient descent, while the parameters of the momentum encoder $g$ are an exponential moving average of the parameters of $f$. Unlike previous MoCo variants~\citep{he2020momentum,chen2020improved}, MoCo-v3~\citep{chen2021empirical} does not utilize a memory queue for storing negative sample representations and replaces the CNN encoder with a ViT~\citep{dosovitskiy2020vit}.

Masked autoencoder (MAE)~\citep{he2022masked} learns representations from unlabeled images by reconstructing the original image based on a partially masked input. Firstly, the input image is divided into patches, followed by randomly masking some of the patches using a uniform distribution. A high masking ratio (75\%) is used in order to design a meaningful pretext task with minimal redundancy. MAE uses an asymmetric architecture where the encoder processes only the visible patches (25\%), thereby significantly reducing the memory and compute requirements. The encoded representations of the visible patches, along with mask tokens for the masked patches, are then processed by a lightweight decoder to reconstruct the original image. Finally, the entire network is trained using a simple reconstruction loss between the original and predicted images, where the loss is computed exclusively on the masked regions:
\begin{equation}
L_{recon}=\texttt{MSE}(\mathbf{x}_M-g(f(\hat{\mathbf{x}}))_M)
\end{equation}
where $f$ and $g$ are the encoder and decoder respectively, and $\mathbf{x}$ and $\hat{\mathbf{x}}$ are the original image and partially masked image input respectively. The subscript $M$ indicates the subset of masked pixels.

\subsection{Fine-tuning Datasets}
\label{sec:finetuning_datasets}
Here, we provide a brief overview of the datasets and evaluation metrics used for each of the 4 downstream tasks.

ChestXray14~\citep{wang2017chestx} is a relatively large dataset comprising 112K chest X-ray scans acquired from 30K patients. Each image in the dataset is associated with classification labels for 14 thoracic diseases. We utilize the official dataset split, with 86K images for training and 25K images for testing. Additionally, we reserve 10\% of the training split for the validation set. Similar to previous work~\citep{haghighi2022dira}, we employ a standard multi-label classification setup and evaluate the classification performance using average AUC across the 14 pathologies.

CheXpert~\citep{irvin2019chexpert} is an open-source dataset comprising 224K chest X-rays obtained from 65K patients. We utilize the official dataset split, with 224K images for training and 234 images for testing. Similar to ChestXray14, we reserve 10\% of the training split for the validation set and use the official test set in order to ensure a fair comparison with CAiD~\citep{taher2022caid} and DiRA~\citep{haghighi2022dira} in Section~\ref{sec:sota}. However, due to the small size of the official CheXpert test set (234 images), we modify the dataset split to create larger validation (11K images) and test (11K images) sets for our fine-tuning analysis experiments in Section~\ref{sec:optimal_ft}. The labels in CheXpert consist of 14 thoracic diseases automatically extracted from radiology reports. However, the official test set is manually labeled by experienced radiologists for 5 diseases including atelectasis, cardiomegaly, consolidation, edema, and pleural effusion. Similar to prior work~\citep{haghighi2022dira}, we follow a standard multi-label classification setup and evaluate the classification performance using the average AUC across the 5 pathologies.

The SIIM-ACR\footnote{https://www.kaggle.com/c/siim-acr-pneumothorax-segmentation/} dataset consists of 10K chest X-ray images along with the corresponding segmentation masks for Pneumothorax. We split the dataset into 70\% for training, 10\% for validation, and 20\% for testing. For the Pneumothorax segmentation task, we train a UNETR~\citep{hatamizadeh2022unetr} model using Dice loss as the training objective, and evaluate the segmentation performance using the mean Dice coefficient.

The NIH Montgomery~\citep{jaeger2014two} dataset consists of 138 chest X-rays, with 80 images classified as normal and 58 images showing Tuberculosis. The dataset also provides segmentation labels for both lungs. Similar to SIIM-ACR, we divide the dataset into 70\% for training, 10\% for validation, and 20\% for testing. We train a UNETR~\citep{hatamizadeh2022unetr} model to perform lung segmentation and evaluate the segmentation performance using the mean Dice coefficient.

\section{Related Work}
\label{sec:appendix_related_work}
In this section, we provide an overview of the relevant work on self-supervised learning and transfer learning.

\subsection{Self-supervised Learning}
Self-supervised learning is a representation learning technique that enables models to learn from unlabeled data without the need for expensive manual annotations. This is achieved via formulating auxiliary tasks where the network is trained to predict certain properties of the data itself. The resulting pre-trained representations can then be utilized for multiple downstream tasks. Various forms of self-supervised learning have emerged in recent years, with contrastive and restorative methods being the most popular.

Contrastive learning~\citep{hadsell2006dimensionality} is a form of self-supervised learning that learns meaningful representations by comparing multiple input samples. The goal is to map the representations of similar images closer together whereas those of dissimilar images farther apart in the embedding space. Instance discrimination~\citep{wu2018unsupervised} is a commonly used form of contrastive learning, where each image is considered as a separate class. Positive (similar) pairs are generated by applying different augmentations to the same image, while negative (dissimilar) pairs are generated by augmenting two distinct images. Contrastive loss functions, such as the InfoNCE loss~\citep{oord2018representation}, are used to train the networks, assigning higher similarities to representations of positive pairs and lower similarities to representations of negative pairs. A wide range of instance discrimination methods have been proposed in recent years~\citep{chen2020simple,he2020momentum,chen2021exploring,chen2020big,zbontar2021barlow}, with the performance strongly dependent on the choice of augmentations~\citep{chen2020simple,grill2020byol}. One drawback of contrastive methods is that they usually need a large number of negative samples in order to learn meaningful representations. However, this reliance on negative samples has been addressed via various solutions such as using large batches~\citep{chen2020simple}, memory banks~\citep{wu2018unsupervised}, and momentum encoders~\citep{he2020momentum}. In this work, we leverage MoCo-v3~\citep{chen2021empirical}, a strong contrastive learning method built upon the vision transformer architecture~\citep{dosovitskiy2020vit}, to develop optimal fine-tuning strategies that effectively utilize contrastive pre-trained features for medical imaging analysis.

Restorative self-supervised learning, also known as masked self-supervised learning, aims to reconstruct missing regions from a partially masked input. There is a long history of restorative pre-training in computer vision~\citep{vincent2010stacked,pathak2016context,dosovitskiy2020vit,he2022masked}. For instance,~\citet{vincent2010stacked} pioneered the use of masking as a type of noise in denoising autoencoders in order to learn higher level representations. Similarly, Context Encoders~\citep{pathak2016context} learnt meaningful features by predicting the pixels of missing regions via CNNs. These pre-trained features have proven to be effective for various downstream tasks including segmentation, classification, and detection. Following the success of restorative (masked) pre-training in NLP~\citep{devlin2018bert}, it has recently gained renewed interest in computer vision as well. A popular restorative pre-training approach is the masked autoencoder (MAE)~\citep{he2022masked}, which employs an asymmetric encoder-decoder architecture and masks 75\% of the image. MAE's self-supervised representations exhibit strong performance, surpassing supervised pre-training in various downstream tasks. In this paper, we leverage the MAE method\citep{he2022masked} to develop optimal fine-tuning strategies that effectively utilize restorative pre-trained features for medical imaging analysis.

Since collecting large labeled datasets is especially difficult in the context of medical imaging, self-supervised learning has proven to be extremely effective for learning representations from unlabeled medical data.~\citet{hosseinzadeh2021systematic} conducted an extensive benchmarking study illustrating the efficacy of several supervised and self-supervised models pre-trained on ImageNet for medical imaging tasks. Additionally, various methods have been proposed for self-supervised learning in medical imaging~\citep{azizi2021big,chaitanya2020contrastive,chen2019self,zhou2021models,tao2020revisiting,haghighi2021transferable,taher2022caid,zhou2021preservational,azizi2022robust,haghighi2022dira}. For instance,~\citet{azizi2022robust} recently introduced REMEDIS, a joint framework that uses both supervised and self-supervised learning for robust and efficient medical imaging, especially in the out-of-distribution setting. Furthermore, some works~\citep{haghighi2021transferable,taher2022caid,zhou2021preservational,haghighi2022dira} have also explored leveraging the complementary benefits of different self-supervised learning strategies. CAiD~\citep{taher2022caid}, for instance, utilizes contrastive and restorative learning in order to learn both global and fine-grained features for medical imaging. DiRA~\citep{haghighi2022dira} builds on top of this by benefiting from adversarial learning as well in addition to contrastive and restorative pre-training. However, despite these advancements in self-supervised \textit{pre-training} algorithms, end-to-end fine-tuning still remains the dominant \textit{fine-tuning} strategy used in several SOTA medical imaging methods~\citep{taher2022caid,haghighi2022dira,azizi2022robust}. In this paper, we design optimal fine-tuning strategies for effectively utilizing self-supervised features for medical imaging, and demonstrate that our best fine-tuning strategies significantly outperform commonly used fine-tuning approaches. Moreover, using the insights from our fine-tuning analysis, we propose a simple yet effective method to directly leverage the complementary features of multiple self-supervised pre-trained models without any need for additional pre-training or delicate hyper-parameter tuning, as in previous methods~\citep{taher2022caid,haghighi2022dira}.
\subsection{Transfer Learning}
Transfer learning~\citep{ge2017borrowing,kumar2018co,chopra2013dlid,chen2015deep} aims to adapt a pre-trained model, trained on a large amount of data, for a downstream task where the data is potentially scarce. Commonly used techniques to utilize the pre-trained knowledge for downstream tasks include fine-tuning the entire network~\citep{girshick2014rich}, updating only the last few layers~\citep{long2015learning}, or simply training a linear classifier on top of a fixed feature extractor~\citep{sharif2014cnn}. A good fine-tuning process should not only learn to solve the downstream task but also retain the pre-trained knowledge in order to avoid overfitting on the downstream task.

To preserve the pre-trained knowledge, various regularization techniques are employed during fine-tuning. For instance,~\citet{xuhong2018explicit} propose an $L^2$ regularizer that encourages similarity between the weights of the fine-tuned model and the pre-trained model, thereby promoting the retention of pre-trained features.~\citet{gouk2021distance} use a distance-based regularisation strategy and constrain the hypothesis class to a small sphere around the pre-trained weights, resulting in improved generalization. Similarly,~\citet{lee2020mixout} introduce mixout, a regularization technique that enhances fine-tuning stability and downstream performance of large language models. Moreover,~\citet{lirethinking} highlight that the optimal hyperparmaters (e.g., effective learning rate, momentum) for fine-tuning tend to be different than those used for training from scratch, which are suboptimal for utilizing the pre-trained knowledge. Moreover,~\citet{jiang2019smart} propose SMART, an efficient computation framework that introduces smoothness-inducing regularization and Bregman proximal point optimization for efficiently fine-tuning large language models, obtaining SOTA performance across several NLP tasks.

In addition to these regularization techniques, freezing specific pre-trained weights has proven to be an effective strategy for retaining prior knowledge and mitigating overfitting. For instance,~\citet{kirkpatrick2017overcoming} address catastrophic forgetting by slowing down learning for certain parameters, thereby retaining knowledge from previous tasks. This approach is shown to achieve SOTA results on several RL problems learnt sequentially.~\citet{ramasesh2020anatomy} study the hidden representations of neural networks and identify deeper layers as the primary source of catastrophic forgetting.~\citet{shen2021partial} observe that not all pre-trained knowledge is necessarily beneficial for few-shot learning tasks, and hence propose to only transfer partial knowledge from the pre-trained model by selectively freezing some layers and fine-tuning others. SpotTune~\citep{guo2019spottune} and AutoFreeze~\citep{liu2021autofreeze} are adaptive fine-tuning approaches that also freeze some layers while adapting others, resulting in improved performance on computer vision and language modeling respectively. Recently,~\citet{lee2022surgical} show that surgical fine-tuning (i.e., fine-tuning only a subset of layers) can yield better results than end-to-end fine-tuning when adapting to various distribution shifts.

Our work aligns more closely with the second line of research. However, unlike existing methods that primarily focus on transferring \textit{supervised} pre-trained models in \textit{natural imaging}, we present the first comprehensive study that discovers effective fine-tuning strategies for \textit{self-supervised} learning in \textit{medical imaging}, where end-to-end fine-tuning is still largely the norm. Since self-supervised methods are trained via quite distinct objectives compared to supervised methods, it is essential to study fine-tuning techniques that effectively leverage self-supervised features for medical imaging. Moreover, considering the various categories of self-supervised learning, with each potentially capturing different types of features, it is not apparent whether the same fine-tuning strategy would be equally effective across all SSL methods. Hence, we design optimal fine-tuning strategies for two popular types of self-supervised learning, namely contrastive and restorative learning, and demonstrate substantial improvements over commonly employed fine-tuning techniques.
\end{document}